\documentclass[runningheads]{llncs}
\usepackage{graphicx}
\usepackage{amsmath}
\usepackage{amssymb}
\usepackage{booktabs}
\usepackage{wrapfig,lipsum}
\usepackage{subcaption}	
\usepackage{xcolor}

\newcommand{\samelineand}{\qquad}

\usepackage[pagebackref,breaklinks,colorlinks]{hyperref}
\usepackage{float}
\usepackage{bm}

\floatstyle{plaintop}
\restylefloat{table}

\begin{document}

\title{Progressive Latent Replay for efficient Generative Rehearsal}

\author{Stanisław Pawlak\inst{1} \and
Filip Szatkowski\inst{1} \and
Michał Bortkiewicz\inst{1} \and \\
Jan Dubiński\inst{1} \and
Tomasz Trzcinski\inst{1,2,3}}
\authorrunning{S. Pawlak et al.}

\institute{$^1$Warsaw University of Technology \samelineand $^2$Jagiellonian University \samelineand $^3$Tooploox \\
\email\{stanislaw.pawlak.dokt, filip.szatkowski.dokt, michal.bortkiewicz.dokt, jan.dubinski.dokt, tomasz.trzcinski\}@pw.edu.pl}
\maketitle          

\begin{abstract}
We introduce a new method for internal replay that modulates the frequency of rehearsal based on the depth of the network. While replay strategies mitigate the effects of catastrophic forgetting in neural networks, recent works on generative replay show that performing the rehearsal only on the deeper layers of the network improves the performance in continual learning. However, the generative approach introduces additional computational overhead, limiting its applications.
Motivated by the observation that earlier layers of neural networks forget less abruptly, we propose to update network layers with varying frequency using intermediate-level features during replay. This reduces the computational burden by omitting computations for both deeper layers of the generator and earlier layers of the main model. We name our method Progressive Latent Replay and show that it outperforms Internal Replay while using significantly fewer resources.

\end{abstract}

\section{Introduction}
\label{sec:intro}
\begin{wrapfigure}{r}{.51\textwidth}
     \vskip  -50pt
            \includegraphics[width=\linewidth]{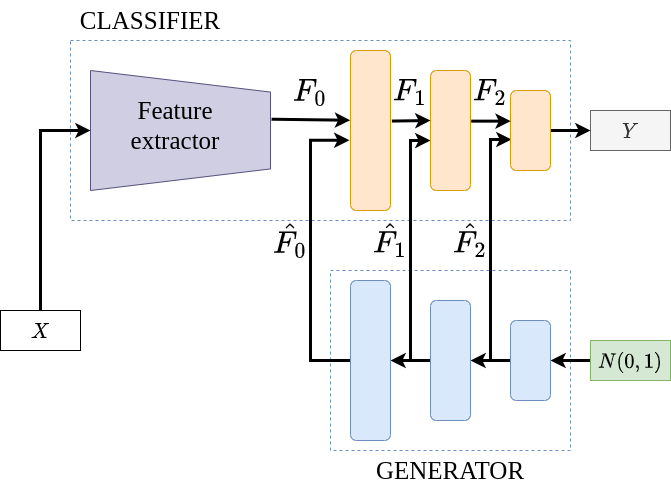}
             \vskip -6pt
     \caption{Overview of the Progressive Latent Replay. We generate features $\hat{F}$ for each layer of the classifier and replay subsets of data at intermediate levels of the network with varying frequencies, which reduces the computational cost of replay.}
     \label{fig:latent_replay}
      \vskip  -0pt
\end{wrapfigure} 

In Continual Learning (CL), we consider a model learning from a stream of tasks. 
One of the main problems in such a scenario is catastrophic forgetting, defined as the decreased performance of the model on the data from the previous tasks when learning something new. 
Forgetting may be alleviated in various ways, including  architecture-based~\cite{yoon2017,Rusu2016}, regularization-based~\cite{kirkpatrick2017,Zenke2017}, and replay-based methods~\cite{Rolnick2019}. 

Although replay-based methods are effective and partially mitigate \clearpage \noindent the forgetting~\cite{1999french}, it comes at the cost of storing past samples in a memory buffer. 
Assuming limited buffer memory,  for high-dimensional data (like images), we can store only a subset of examples for replay. 

Recent works introduce Internal Replay (IR) to increase the efficiency of the replay \cite{merlinPracticalRecommendationsReplaybased2022,van2020brain}. The main idea of IR is to rehearse representations from internal layers of the network instead of full-size images. IR improves rehearsal efficiency and performance on more complex datasets~\cite{van2020brain}, especially in a generative approach (i.e. using an additional generative model to produce replay data online during training instead of storing it in a buffer ~\cite{shin2017continual}).  Following IR, we investigate this approach's limitations as well as performance and resource utilization trade-offs. We show that pretraining of the early network layers is essential for IR and demonstrate that naive IR fails without pretraining. Motivated by the use of pretraining in CL~\cite{mehtaEmpiricalInvestigationRole2021,van2020brain}, showing that initial network layers should not change significantly during continual learning training, and inspired by recent works on forgetting in neural networks, which state that forgetting occurs mainly in the final layers of the network\cite{ramaseshAnatomyCatastrophicForgetting2020}  we propose to further improve the training by changing the frequency of updates for layers of the main model during the replay phase.

The main contribution of this work is the proposed Progressive Latent Replay method, a generalized version of Internal Replay, which reduces the computational cost of rehearsal and improves the overall model performance by replaying internal representations on different levels of the base network. The representations from the deeper layers are repeated more frequently, while the representations from the earlier ones less frequently. Replay with our method is more efficient, as we can sample intermediate-level features, omitting computations for both deeper layers of the generator and earlier layers of the main model.

\section{Related Work}
Continual Learning methods can be divided into three main categories~\cite{mundt2020wholistic}. \newline Architecture-based methods~\cite{yoon2017,Rusu2016} work by assuming that the neural network structure is dynamic and can be modified during training to mitigate catastrophic forgetting. Regularization-based methods~\cite{kirkpatrick2017,Zenke2017} approach this problem by regularizing the network during training.
Replay methods~\cite{Rolnick2019,shin2017continual} aim to prevent catastrophic forgetting by rehearsing some examples from previous tasks along with the examples belonging to the current task. 
\paragraph{Replay in Continual Learning}
The simple replay approach stores examples in a memory buffer and update the buffer with every new task~\cite{mundt2020wholistic}. However, extending the buffer with new data becomes inefficient with an increasing number of tasks. 
A possible solution is using example selection methods~\cite{mundt2020wholistic}. Yet, it is not trivial, because we do not know beforehand what examples would be most valuable for replay in the following tasks.

To address the above-mentioned issues, in~\cite{shin2017continual} authors proposed generative replay where the generative model is trained to model the distribution of data from previous tasks. Rehearsal samples produced by the generator are interleaved with new data during training of the base model. Generative replay framework is often based on different types generative models, like Variational Autoencoders~\cite{kemkerFearNetBrainInspiredModel2018} %HyperNetworks~\cite{vonoswaldContinualLearningHypernetworks2019}
or GANs~\cite{zhaiLifelongGANContinual2019}.

 \paragraph{Internal Replay}
 In \cite{ramaseshAnatomyCatastrophicForgetting2020} authors show that the forgetting is not equal for all network layers and depends highly on the depth of the layer. The first layers of the network act as feature extractors and forget slowly, while the deeper layers are most prone to forgetting as they need to adapt the most to accommodate new tasks. Thus, the idea of rehearsing network internal representations during replay called Internal Replay\footnote{Internal Replay approach is also called latent~\cite{pellegriniLatentReplayRealTime2020a} or feature~\cite{Thandiackal2021} replay} (IR) gained much interest in recent works connected with generative~\cite{van2020brain,Thandiackal2021,kemker2018fearnet,xiang2019incremental} and buffer replay~\cite{pellegriniLatentReplayRealTime2020a}. While in the standard replay, data is propagated through all network layers during rehearsal, Internal Replay rehearses the data only for deeper layers. 
 
 Authors of Brain-Inspired Replay~(BI-R)~\cite{van2020brain} propose to use pretrained first network layers as a fixed feature extractor and perform replay only on hidden fully-connected layers. Moreover, they show in the ablation study that Internal Replay modification yields the greatest performance improvement. We build upon and extend this work by introducing internal replay with varying frequencies with respect to network layer depth.
 
 In \cite{pellegriniLatentReplayRealTime2020a} authors investigate another approach related to internal representations of the model. In contrast to BI-R, their work is based on a storage buffer. In the proposed method, named Latent Replay, the authors explore the impact of replaying latent patterns obtained from the particular hidden layer. Specifically, instead of storing the data in the buffer, they store intermediate representations obtained from incoming data and later use them for replay. While they store representations of only one internal layer, we propose using a generative model to produce generations on multiple internal layers.
 
 Works connected with GANs \cite{Thandiackal2021,xiang2019incremental} use internal representations to improve or condition the training of a generative model. Additionally, in \cite{Thandiackal2021} authors stress the necessity for a custom pretraining as a major flaw of other approaches leveraging Internal Replay. Motivated by this, we investigate the impact of pretraining in BI-R, showing that it is essential for Internal Replay to work properly.

\section{Method}

\begin{figure}
    \centering
    \includegraphics[width=0.8\linewidth]{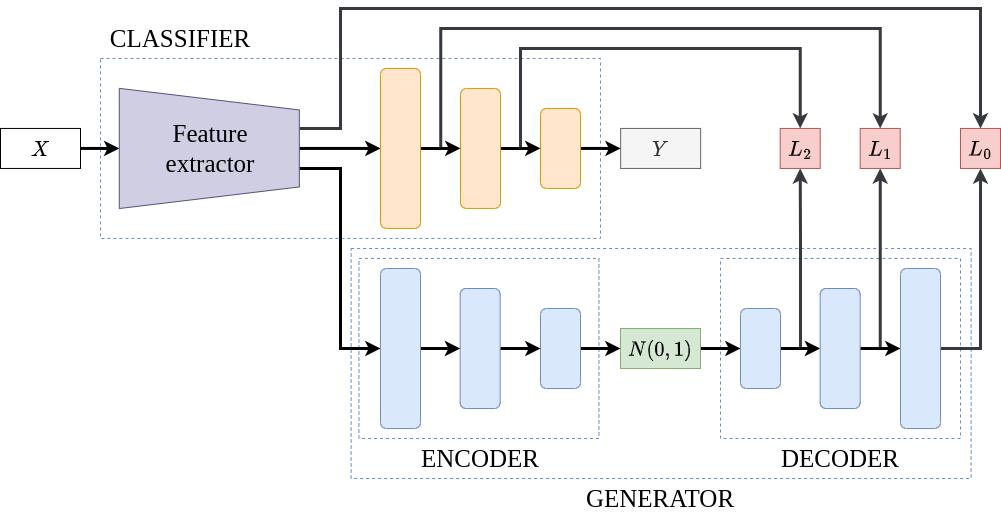}
    \caption{Generative model training in Progressive Latent Replay. We define the architecture of the generative model so that input and output dimensions of corresponding encoder layers are identical to fully-connected counterparts of the classifier. The decoder layers are defined so that their shapes match the reversed shapes of the encoder. Then we optimize the generative model to reconstruct features encoded by the classifier at every intermediate layer.}
    \label{fig:latent_training}
\end{figure}

In this section, we present our Progressive Latent Replay (PLR) method for rehearsal in a continual learning classification problem. PLR performs replay on different layers of the main model (i.e. the classifier) with varying frequencies.
We also describe update strategies and the relative update cost metric used to measure the efficiency of our method.

\subsection{Progressive Latent Replay}\label{PLR}
We propose to update deeper layers more frequently than the earlier ones to reduce the computational cost of replay. Thus, we generalize the Internal Replay proposed in BI-R~\cite{van2020brain} to Progressive Latent Replay, where the updates can be done on multiple levels of the network with varying frequencies. Replaying low-level features less frequently results in fewer weight updates at earlier layers, saving computational resources. During the replay phase, we sample different-level features at intermediate layers of the generator and replay them at corresponding levels of the classifier, as shown in Figure~\ref{fig:latent_replay}. The modifications required to train the generator to produce intermediate-level features are presented in Figure~\ref{fig:latent_training}.

We use a model based on variational autoencoder \cite{kingma203vae} as a generator, but introduce slight modifications to be able to replay intermediate-level features. Typically, variational autoencoders are trained by minimizing the following loss function: 
\begin{equation}
    L_{vae} =  L_{recon} +  L_{latent} = E_{q_{\theta}(z \mid x)}\left[\log p_{\varphi}(x \mid z)\right]-\operatorname{KL}\left(q_{\theta}(z \mid x) \| p(z)\right)
\end{equation}
where $q_{\theta}$, $p_{\varphi}$ are the encoder and the decoder networks, $x$ is the input data, $z$ is the sampled latent vector and KL denotes the Kullback–Leibler divergence.

In our method we use the sampling strategy and latent regularization loss $L_{latent}$ from \cite{van2020brain}. Moreover, for an effective rehearsal of intermediate features, we introduce an alternative reconstruction loss $L_{recon}$ as the generative model must generate features of sufficient quality not only in the end-to-end fashion but also in every corresponding decoder layer. We obtain this by defining the reconstruction loss of the variational autoencoder $L_{recon}$ as a sum of losses for features generated for each network layer:
\begin{equation}
    L_{recon} = \sum_{n=0}^{N-1} L(F_{n}, \hat{F_{n}}),  
\end{equation}
where $F_{n}$ and  $\hat{F_{n}}$ stand for latent features obtained by propagating real data through the classifier up to the layer $n$ and reconstructions generated by autoencoder for this layer. $L$ denotes the reconstruction loss for individual layers, which in our method is mean squared error.

\begin{figure}[!t]
    \centering
    \includegraphics[width=0.8\linewidth]{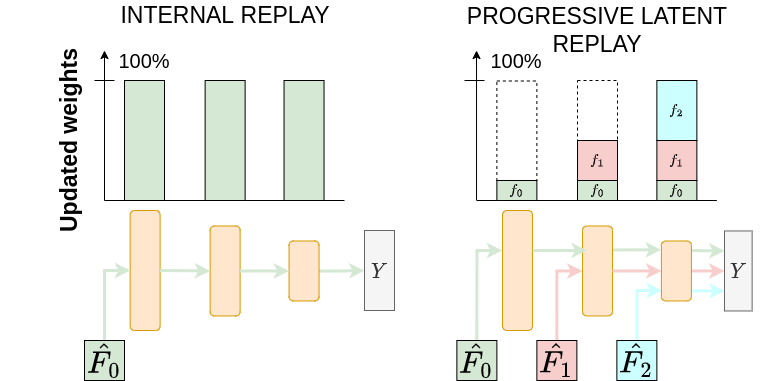}
    \caption{Comparison of Internal Replay and Progressive Latent Replay with strategy $S=[f_0, f_1, f_2]$. Replaying low-level features less frequently results in fewer weight updates at earlier layers, saving computational resources.}
    \label{fig:latent_strategies}
\end{figure}

The generator trained with modified reconstruction loss allows us to sample and replay intermediate-level features with any frequency, omitting computations for both deeper layers of the generator and earlier layers of the main model.  We define layer update frequency $f_n$ for $n$th layer of the classifier as the fraction of the replay data generated at this level for a single training step, so:
\begin{equation}
    \sum_{k=0}^{N-1}f_{k} = 1.
\end{equation}

Then we can denote Progressive Latent Replay update strategy for the network with $N$ layers as $S = [f_{0}, f_{1}, ..., f_{N-1}]$. We compare our method with Internal Replay in Figure~\ref{fig:latent_strategies}.

In comparison with traditional replay methods, our method updates only a small fraction of the weights, while achieving similar or better results. The only increase in the computational cost of training the generator for Progressive Latent Replay comes from modifications to the reconstruction loss function and is negligible. We describe the computational cost of our method in detail in the following section.

\subsection{Computational efficiency of the method}

We measure the efficiency of Progressive Latent Replay strategies by comparison with Internal Replay. We estimate $U(S)$ - the number of parameters updated for strategy $S=[f_0, ..., f_{N-1}]$ - as follows:
\begin{equation}
    U(S) = U([f_0, ..., f_{N-1}]) = \sum_{n=0}^{N-1} \sum_{k=0}^{n}f_{k} P_{n},
\end{equation}
where $P_{n}$ stands for the total number of weights in $n$th layer. We estimate the number of weights updated in Internal Replay using the same formula as $U([1, 0, ..., 0])$, because Internal Replay can be be described as a special case of Progressive Latent Replay where we always update all the layers.
We define the relative cost of our method $R$ as:
\begin{equation}\label{eq:R}
    R(S) = \frac{U(S)}{U([1, 0, ..., 0])}.
\end{equation}
 $U([1, 0, ..., 0])$ is the highest possible number of updates and equals the number of updates performed during Internal Replay. For for any Progressive Latent Replay strategy $S$:
 \begin{equation}
     0 < R(S) <= 1.
 \end{equation}

This metric allows us to measure how many weights are updated during the replay, which directly corresponds with the resource consumption as sampling and replaying the features at deeper layers of the network requires fewer computations. Lower values of $R(S)$ mean that fewer updates are performed at the earlier layers of the network, which does not require propagating information through all the layers in the generator and main model. With a correct replay strategy, our method can significantly reduce the number of computations during the replay, while achieving similar or even better results.

\section{Experiments}

Our experiments focus on comparing our Progressive Latent Replay with Internal Replay using performance and efficiency measures. To strengthen the motivation for our method we first evaluate the impact of pretraining and freezing of initial network layers on model performance in a continual learning setup. Then, we show on a simple two-tasks setup with rehearsal of internal representations from a buffer, that more frequent replay on deeper layers of the base model lowers the number of updates needed during CL training with almost no performance decrease. Finally, we perform a full comparison between our PLR and IR on a much more challenging setup with ten tasks on split-CIFAR100 using generative replay approach. 

\subsection{Experimental Setup}
We define generator -- the autoencoder -- so that the encoder architecture is identical to the stack of fully-connected layers as in the classifier. The decoder then mirrors encoder architecture, so input dimensionality of encoder layers and fully-connected layers in the main model matches output dimensionality of corresponding layers in decoder. In all cases, replay is done with features from fully-connected layers, and we do not perform any replay for the feature extractor. We illustrate the training of the generator in Figure~\ref{fig:latent_training}. 

 We conduct our experiments on three datasets:  CIFAR10, CIFAR100 and FashionMNIST, and perform training in a class incremental (CI) scenario~\cite{van2019three}. Thus, datasets are split sequentially into tasks. We obtain two tasks with five classes for CIFAR10, ten tasks with ten classes in the case of CIFAR100 and five tasks with two classes for FashionMNIST. The batch size is set to 256 in all experiments.

\subsection{Model architecture}
We follow BI-R~\cite{van2020brain} in architecture design for Progressive Latent Replay evaluation. The main model is constructed from a feature extractor and a stack of fully-connected layers. We use the modified autoencoder described in Section~\ref{PLR} as the generator and define the encoder architecture so that it is identical to the stack of fully-connected layers in the main model. The decoder then mirrors encoder architecture, so the input dimensionality of encoder layers and fully-connected layers in the main model matches the output dimensionality of corresponding layers in the decoder. In all cases, replay is done with features from fully-connected layers, and we do not perform any replay for the feature extractor. We illustrate the training of the generator in Figure~\ref{fig:latent_training}. 
    
For experiments on CIFAR10/100, the model consists of five convolutional layers as a feature extractor followed by fully-connected layers. We use ReLU as an activation function, and the model output consists of a Softmax layer. We use two different configurations of the fully-connected layers: \textbf{ARCH1} with two fully-connected layers with 2000 units followed by a final layer with 100 neurons, and \textbf{ARCH2} with three fully-connected layers with 1000 units followed by a final layer with 100 neurons. The first configuration mirrors BI-R, while the second consists of more layers so that we can examine the impact of update strategies in more detail.

For additional experiments on FashionMNIST, the model has three convolutional layers followed by three fully-connected layers with 50 units each and a final Softmax output layer.

\begin{figure}[!b]
\centering
  \begin{minipage}[t]{0.48\textwidth}
    \begin{flushleft}
          \centering
          \vskip -20pt
          \includegraphics[width=1\linewidth]{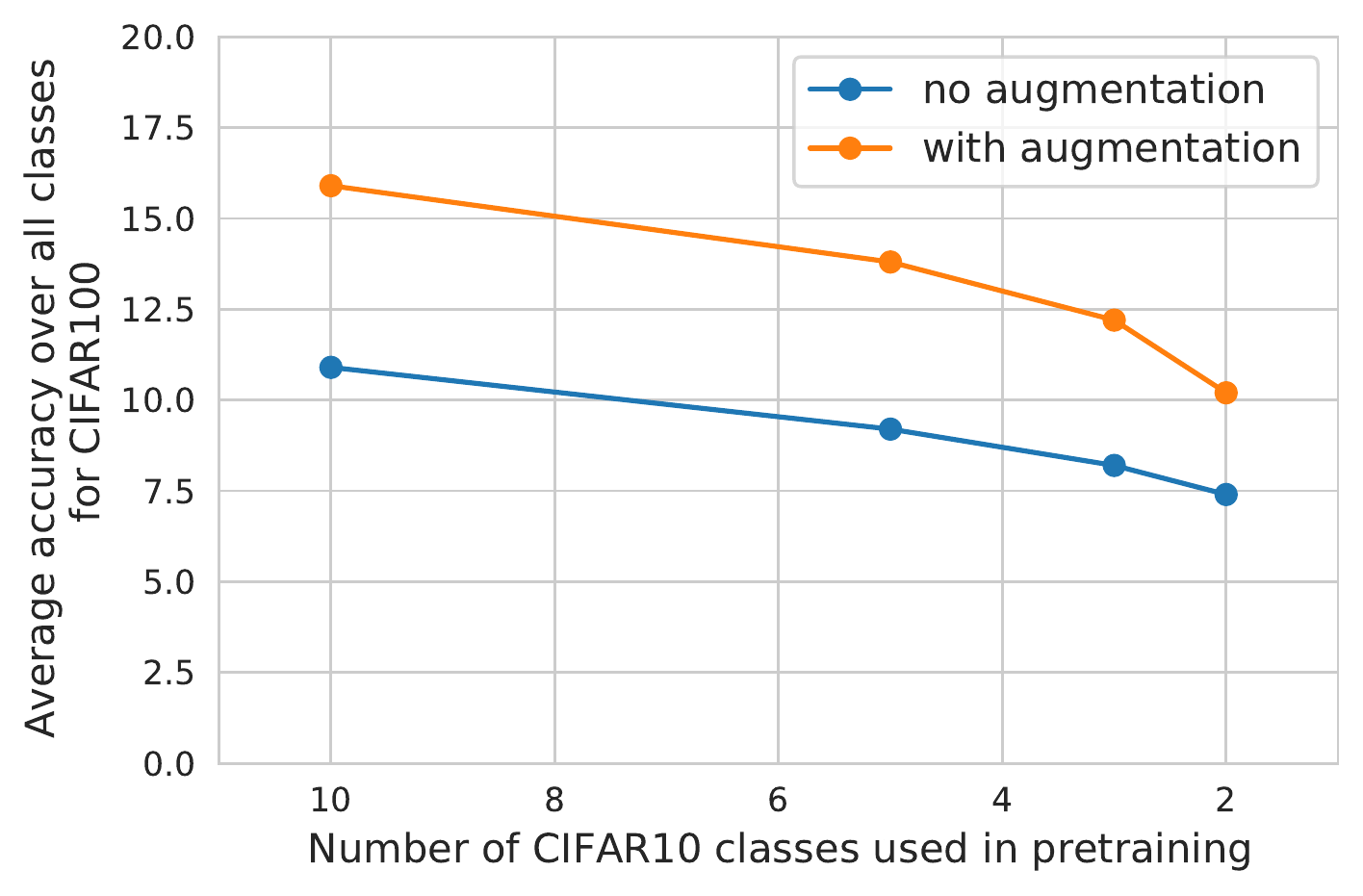}
          \vskip -6pt
          \caption{Effects of limiting the pretraining dataset on the quality of Internal Replay. We change the number of classes from CIFAR10 on which we train the feature extractor before evaluating the performance of Internal Replay.}
          \label{fig:ablation_pretrain}
    \end{flushleft}
\end{minipage}%
 \begin{minipage}[t]{0.04\textwidth}
    \begin{flushleft}
        \centering
    \end{flushleft}
\end{minipage}%
\begin{minipage}[t]{0.48\textwidth}
    \begin{flushright}
    \vskip -20pt
        \includegraphics[width=1\linewidth]{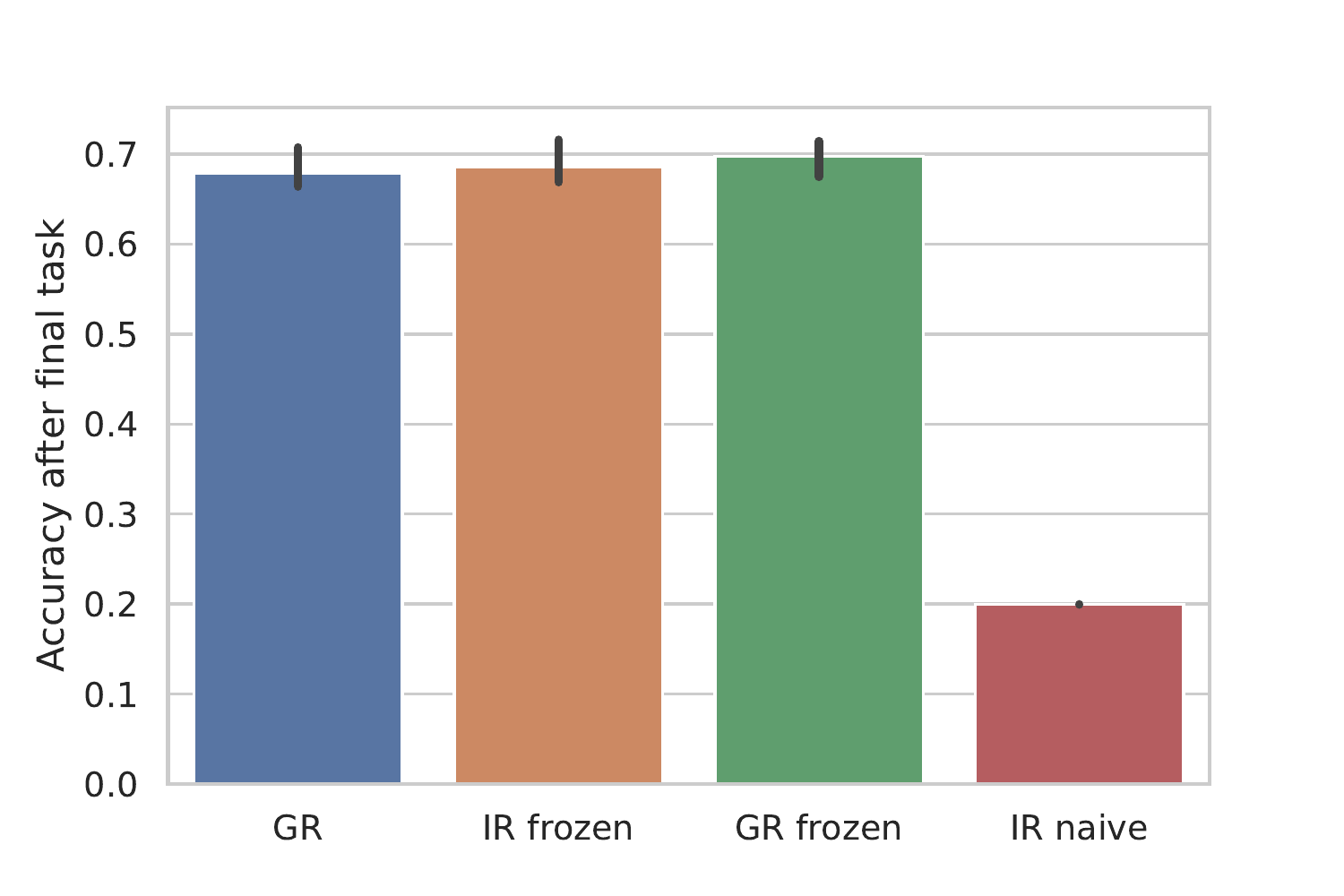}
        \vskip -6pt
        \caption{Custom pretraining alternatives for simple datasets. We report average accuracy after the final task in class incremental setup on FashionMNIST for four different setups: standard generative replay (GR), Internal Replay (IR) with encoder part frozen after 1st task, GR with encoder and decoder part frozen after 1st task, and IR with no freezing (naive).}
        \label{fig:gr_pretrain_strategies}
    \end{flushright}
\end{minipage}%
\end{figure}

To evaluate Progressive Latent Replay strategies on CIFAR100, encoder weights were pretrained on CIFAR10, following the base setup introduced in BI-R. Further ablation study on pretraining is presented in Figure \ref{fig:ablation_pretrain}. 

\subsection{Metrics} 
We use Frechet Inception Distance (FID) metric to evaluate the quality of generations produced by the generative model. This metric originally accepts only full-size images as input and compares distributions of internal
representations from the inception model pretrained on the external dataset. Thus, we use the modified version introduced in~\cite{van2020brain} to evaluate internal representations of a different size than in the inception network. For modified FID calculation, we change the reference model from inception to a model pretrained on the joined training dataset, which has the same base architecture as the one evaluated.
  
\subsection{Pretraining impact on Internal Replay}

In Figure \ref{fig:gr_pretrain_strategies} we show that the combination of pretraining and freezing the encoder layers is required for Internal Replay to work correctly. In other words, the naive approach using Internal Replay with no pretraining and no freezing layers fails even on a FashionMNIST, which is much simpler than CIFAR100.
An alternative approach to pretraining on an external dataset is freezing the weights in the encoder part of the network after the first task. We show that such an approach combined with IR obtains results comparable to standard generative replay but uses fewer resources for replay updates.
Moreover, we evaluate how the performance of IR depends on the dataset used for pretraining (see Figure~\ref{fig:ablation_pretrain}). More specifically, we train a classifier with convolutional layers on a different number of classes from CIFAR10. Then we use the convolutional part of the network as a feature extractor. We evaluate two scenarios: with and without data augmentation. Results presented in Figure~\ref{fig:ablation_pretrain} clearly show the importance of pretraining with data augmentation for Internal Replay, as the achieved accuracy drops significantly for a lower number of classes in the dataset and when pretraining is not used.
These insights (i.e. that the pretraining is essential for IR and additional updates on early layers reduce performance and efficiency of IR) motivated us to propose a depth-dependent frequency of updates during replay to increase efficiency even further by omitting some of the earlier layers updates.

\begin{table}[t!]
\centering
 \begin{tabular}{||c|c|c|c||} 
  \hline
 Architecture & Strategy & R $\downarrow$ & Accuracy $\uparrow$  \\ [0.5ex] 
 \hline\hline
 ARCH1 & Internal Replay & 100\% &  \textbf{71.2\%} \bm{$\pm$} \textbf{0.7}\%\\ 
 &  S=[0.7, 0.3]  & 71.4\%  & 71.1\% $\pm$ 0.7\% \\
 & S=[0.5, 0.5]  & 52.4\%  &  71.0\% $\pm$ 0.8\% \\
  & S=[0.3, 0.7]  & \textbf{33.3}\%  & 70.6\% $\pm$ 1.0\% \\
 \hline\hline
 ARCH2 & Internal Replay & 100\% &  \textbf{70.2\%} \bm{$\pm$} \textbf{0.7}\%\\ 
  & S=[0.5,0.3,0.2]  & 66.7\%  &   69.9\% $\pm$ 0.6\%\\
  & S=[0.34,0.33,0.33] & 52.9\%  &  69.3\% $\pm$ 0.7\% \\
  & S=[0.2,0.3,0.5]  & \textbf{38.1}\% &   69.2\% $\pm$ 0.8\% \\
 \hline
\end{tabular}
\caption{Updating deeper layers with higher frequency during replay results in similar performance to IR with significantly less replay updates. We report test accuracy $\pm$ SEM after second task for split-CIFAR10 class-incremental scenario with buffer replay containing 512 internal representations of samples. Results are averaged between three runs with different seeds.}
 \label{tab:base_plr_example}
\end{table}
  
\subsection{Progressive Latent Replay with different strategies}
In Table \ref{tab:base_plr_example} we show that for a simple continual learning setup (only two tasks, each containing five classes) omitting some replay updates on earlier layers results in a similar performance to IR. At the same time, each strategy performs much fewer replay updates compared to IR (see $R$ metric). For simplicity, in this experiment we use a replay buffer with internal representations for each hidden layer instead of an additional generative model (for both IR and PLR-based strategies).

A full comparison of Progressive Latent Replay and Internal Replay we perform using a more challenging setup with ten tasks on split-CIFAR100. We show results obtained with IR and different Progressive Latent Replay update strategies for architectures ARCH1 and ARCH2 in Table \ref{tab:fc3_lr}. We compare regular IR with three Progressive Latent Replay strategies in terms of Relative Cost (R~--~see Equation \ref{eq:R}), modified FID (mFID), and test accuracy averaged over all tasks. We choose strategies that vary considerably in terms of update frequencies per layer to explore the Progressive Latent Replay properties more closely.

We find that Progressive Latent Replay achieves significantly better performance in terms of accuracy for all strategies while using less computational resources than Internal Replay. 
The results are consistent across the two architectures, with the best strategies updating the first layer only half of the time. 
 
The increased final performance in terms of accuracy for PLR strategies in the generative approach shows that training the generator with additional constraints on intermediate features results in better features produced for replay. It is a promising premise for future work on scaling generative replay with progressive latent replay updates to longer sequences of tasks in continual learning. 

\begin{table}[t!]
\centering
 \begin{tabular}{||c|c||c|c|c||} 
  \hline
 Architecture & Strategy & R $\downarrow$ & mFID $\downarrow$ & Accuracy $\uparrow$  \\ [0.5ex] 
 \hline\hline
 ARCH1 & Internal Replay & 100\% &  169 & 15.9\% $\pm$ 0.7\% \\ 
 &  S=[0.7, 0.3]  & 71.4\%  & 142  & 19.9\%  $\pm$ 0.9\% \\
 & S=[0.5, 0.5]  & 52.4\%  & 144 & \textbf{20.3\%}  \bm{$\pm$} \textbf{0.9\%} \\
  & S=[0.3, 0.7]  & \textbf{33.3}\%  &  \textbf{140}  & 20.1\% $\pm$ 1\% \\
 \hline\hline
 ARCH2 & Internal Replay & 100\% &  514  & 11.3\% $\pm$ 0.8\% \\
  & S=[0.5,0.3,0.2]  & 66.7\%  & 869 & \textbf{14.5}\% \bm{$\pm$} \textbf{0.9\%} \\
  & S=[0.34,0.33,0.33] & 52.9\% & \textbf{318} & 13.5\% $\pm$ 0.7\% \\
  & S=[0.2,0.3,0.5]  & \textbf{38.1}\% & 329  & 13.4\% $\pm$ 0.5\% \\
 \hline
\end{tabular}
 \caption{ Results in class-incremental scenario on split-CIFAR100 dataset averaged from three runs with different seeds. We compare Internal Replay with different Progressive Latent Replay update strategies for ARCH1 and ARCH2. We report test accuracy $\pm$ SEM after the final task.}
 \label{tab:fc3_lr}
\end{table}

\section{Conclusion}
In this paper, motivated by the observation that forgetting occurs mainly in the final layers of neural networks, we introduce Progressive Latent Replay, a generalization of Internal Replay. We train the generator to produce features for intermediate layers of the main model and replay the different level features with varying frequencies. We show that Progressive Latent Replay improves on standard Internal Replay while using significantly fewer resources for replay. 

Our work is a step toward efficient continual learning in scenarios where computational resources are constrained, or access to pretraining data is restricted. We show that performing weight updates with varying frequency for each layer improves the efficiency of replay while preserving similar or even increasing the overall model performance, which confirms that layers in neural networks forget differently. In future works, we would like to study this phenomenon in more detail and determine the best update strategies for Progressive Latent Replay.

{\small
\bibliographystyle{splncs04}
\bibliography{egbib}

\begin{thebibliography}{10}
\providecommand{\url}[1]{\texttt{#1}}
\providecommand{\urlprefix}{URL }
\providecommand{\doi}[1]{https://doi.org/#1}

\bibitem{Rolnick2019}
{D. Rolnick et al.}: Experience {R}eplay for {C}ontinual {L}earning. In:
  NeurIPS (2019)

\bibitem{1999french}
French, R.M.: Catastrophic forgetting in connectionist networks. Trends in cog.
  scie.  (1999)

\bibitem{kemkerFearNetBrainInspiredModel2018}
Kemker, R., Kanan, C.: {{FearNet}}: {{Brain-Inspired Model}} for {{Incremental
  Learning}} (Feb 2018)

\bibitem{kemker2018fearnet}
Kemker, R., Kanan, C.: Fearnet: Brain-inspired model for incremental learning.
  In: International Conference on Learning Representations (2018),
  \url{https://openreview.net/forum?id=SJ1Xmf-Rb}

\bibitem{kingma203vae}
Kingma, D.P., Welling, M.: Auto-encoding variational bayes (2013).
  \doi{10.48550/ARXIV.1312.6114}, \url{https://arxiv.org/abs/1312.6114}

\bibitem{kirkpatrick2017}
Kirkpatrick, J., Pascanu, R., Rabinowitz, N., Veness, J., Desjardins, G., Rusu,
  A.A., Milan, K., Quan, J., Ramalho, T., Grabska-Barwinska, A., Hassabis, D.,
  Clopath, C., Kumaran, D., Hadsell, R.: Overcoming catastrophic forgetting in
  neural networks. PNAS  (2017)

\bibitem{mehtaEmpiricalInvestigationRole2021}
Mehta, S.V., Patil, D., Chandar, S., Strubell, E.: An {{Empirical
  Investigation}} of the {{Role}} of {{Pre-training}} in {{Lifelong Learning}}
  (Dec 2021)

\bibitem{merlinPracticalRecommendationsReplaybased2022}
Merlin, G., Lomonaco, V., Cossu, A., Carta, A., Bacciu, D.: Practical
  {{Recommendations}} for {{Replay-based Continual Learning Methods}}.
  arXiv:2203.10317 [cs]  (Mar 2022)

\bibitem{mundt2020wholistic}
Mundt, M., Hong, Y.W., Pliushch, I., Ramesh, V.: A wholistic view of continual
  learning with deep neural networks: Forgotten lessons and the bridge to
  active and open world learning (2020)

\bibitem{pellegriniLatentReplayRealTime2020a}
Pellegrini, L., Graffieti, G., Lomonaco, V., Maltoni, D.: Latent {{Replay}} for
  {{Real-Time Continual Learning}} (Mar 2020)

\bibitem{ramaseshAnatomyCatastrophicForgetting2020}
Ramasesh, V.V., Dyer, E., Raghu, M.: Anatomy of {{Catastrophic Forgetting}}:
  {{Hidden Representations}} and {{Task Semantics}}. arXiv:2007.07400 [cs,
  stat]  (Jul 2020)

\bibitem{Rusu2016}
Rusu, A.A., Rabinowitz, N.C., Desjardins, G., Soyer, H., Kirkpatrick, J.,
  Kavukcuoglu, K., Pascanu, R., Hadsell, R.: Progressive {N}eural {N}etworks
  (2016), arXiv:1606.04671

\bibitem{shin2017continual}
Shin, H., Lee, J.K., Kim, J., Kim, J.: Continual learning with deep generative
  replay (2017)

\bibitem{Thandiackal2021}
Thandiackal, K., Portenier, T., Giovannini, A., Gabrani, M., Goksel, O.: Match
  what matters: Generative implicit feature replay for continual learning. CoRR
   \textbf{abs/2106.05350} (2021), \url{https://arxiv.org/abs/2106.05350}

\bibitem{van2020brain}
van~de Ven, G.M., Siegelmann, H.T., Tolias, A.S.: Brain-inspired replay for
  continual learning with artificial neural networks. Nature communications
  \textbf{11}(1),  1--14 (2020)

\bibitem{van2019three}
Van~de Ven, G.M., Tolias, A.S.: Three scenarios for continual learning. arXiv
  preprint arXiv:1904.07734  (2019)

\bibitem{xiang2019incremental}
Xiang, Y., Fu, Y., Ji, P., Huang, H.: Incremental learning using conditional
  adversarial networks. In: Proceedings of the IEEE/CVF International
  Conference on Computer Vision. pp. 6619--6628 (2019)

\bibitem{yoon2017}
Yoon, J., Yang, E., Lee, J., Hwang, S.J.: Lifelong {L}earning with
  {D}ynamically {E}xpandable {N}etworks. In: ICLR (2018)

\bibitem{Zenke2017}
Zenke, F., Poole, B., Ganguli, S.: Continual learning through synaptic
  intelligence. In: Proceedings of the 34th International Conference on Machine
  Learning - Volume 70. p. 3987–3995. ICML'17, JMLR.org (2017)

\bibitem{zhaiLifelongGANContinual2019}
Zhai, M., Chen, L., Tung, F., He, J., Nawhal, M., Mori, G.: Lifelong {{GAN}}:
  {{Continual Learning}} for {{Conditional Image Generation}}. In: 2019
  {{IEEE}}/{{CVF International Conference}} on {{Computer Vision}} ({{ICCV}}).
  pp. 2759--2768. {IEEE}, {Seoul, Korea (South)} (Oct 2019).
  \doi{10.1109/ICCV.2019.00285}

\end{thebibliography}
}
\end{document}